\documentclass[10pt,twocolumn,letterpaper]{article}

\usepackage{cvpr}
\usepackage{times}
\usepackage{epsfig}
\usepackage{graphicx}
\usepackage{amsmath}
\usepackage{amssymb}
\usepackage{multirow}
\usepackage{verbatim}

\usepackage[breaklinks=true,bookmarks=false]{hyperref}

\cvprfinalcopy 

\def\cvprPaperID{****} 

\begin{document}

\title{Towards computer vision powered color-nutrient assessment of pur\' eed food}

\author{Kaylen J. Pfisterer$^{*,1,2}$, \text{Robert Amelard}$^{2}$, \text{Braeden Syrnyk}$^{1}$,\text{ and Alexander Wong}$^{1,2}$\\
$^{1}$ Vision and Image Processing Research Group, University of Waterloo, Waterloo, ON, Canada \\
$^{2}$ Schlegel-UW Research Institute for Aging, Waterloo, ON, Canada \\
$^{*}${\tt\small kpfisterer@uwaterloo.ca}
}
\maketitle

\begin{abstract}
With one in four individuals afflicted with malnutrition, computer vision may provide a way of introducing a new level of automation in the nutrition field to reliably monitor food and nutrient intake. In this study, we present a novel approach to modeling the link between color and vitamin A content using transmittance imaging of a pur\' eed foods dilution series in a computer vision powered  nutrient sensing system via a fine-tuned deep autoencoder network, which in this case was trained to predict the relative concentration of sweet potato pur\' ees. Experimental results show the deep autoencoder network can achieve an accuracy of 80\% across beginner (6 month) and intermediate (8 month) commercially prepared pur\' eed sweet potato samples. Prediction errors may be explained by fundamental differences in optical properties which are further discussed.
\end{abstract}

\section{Introduction}

 Worldwide, approximately 590 million people are afflicted with dysphagia~\cite{cichero2017}, placing these indviduals at increased risk for malnutrition~\cite{sura2012,ilhamto2014}. Pur\' eed foods can allow for safe ingestion however nutritional composition can be highly variable from preparation method differences~\cite{ilhamto2014}. Recently established standards and guidelines assist with assessing quality based on visual and mechanical tests~\cite{cichero2017} however they involve time-intensive manual tasks. Image-based assessment may enable a new layer of automation and provide an objective tool for assessing pur\' eed food quality as they are well modeled as homogeneous media, and are the most relevant to individuals living with dysphagia.
 
 In this study, we focus here on our work on vitamin A, a family of chromophores including carotenoids that are optically active in the visible spectrum~\cite{melendez2007}. This builds on our previous work predicting pur\' ee relative nutritional density with deep learning in a small data set~\cite{pfisterer2018} and exploration of nutrient-color link~\cite{pfisterer2018differential}. Here, we significantly expand our preliminary vitamin A assessment results. The rest of this paper is organized into: research methods and data acquisition, high-level results and an in-depth discussion of vitamin A composition in sweet potato and carrot. Finally, we discuss future directions in the context of computer vision powered biophotonics.
 

\section{Methods}
We have developed a two pronged end-to-end system for tracking LTC resident food and fluid intake which co-integrates machine learning and computer vision with biophotonic analysis. Figure~\ref{fig:Lab} shows a graphical representation of the system architecture. Fine-grained, single nutrient assessment is accomplished through biophotonic analysis and yields a CIELAB Gaussian distribution for each flavor and dilution, and a \% transmittance map comparing highest (sweet potato) and lowest (carrot) vitamin A content. A five-tier dilution series relative to the initial concentration for each of five commercially-prepared pur\' eed foods containing vitamin~A: butternut squash, carrot, mango, and 6 and 8 month sweet potato. Thirty fullfield white normalized transmittance images were acquired using 15 mL samples in petri dishes with a broadband tungsten-halogen light source and front glass fabric diffuser under a glass loading plate. Pixelwise spectral transmittance was computed using white and dark normalization:
\begin{equation}
    T(x,y) = \frac{I(x,y) - I_d(x,y)}{I_w(x,y) - I_d(x,y)} \frac{\tau_2}{\tau_1}
\end{equation}
where $I$, $I_d$, and $I_w$ are the normal, dark, and white images respectively, and $\tau_1$, $\tau_2$ are the exposure times during normal and white image acquisition respectively. To avoid photon boundary artifacts, a 35$\times$35~mm region in the center of the sample was used to analyze the distribution of pixel transmittance spectra. Color values were converted to CIELAB color space, which accurately captures the underlying chemical structure (i.e., conjugated double bonds) of carotenoid variants~\cite{melendez2007}. Images were processed using a photon migration model in a homogeneous pur\' ee mixture and Beer-Lambert exponential decay of light attenuation~\cite{pfisterer2018differential}:
\begin{equation}
T = \frac{I}{I_0} = \exp(\varepsilon_{H_2O} \cdot c_{H_2O} \cdot l_{H_2O} + \varepsilon_{vitA} \cdot c_{vitA} \cdot l_{p})
\end{equation}
where $T$ is transmittance, $I_0$ and $I$ are the incident and transmitted illumination respectively. In the selected pur\' ees, we assume that the dominant absorber is vitamin A. For ${H_2O}$ and ${vitA}$, $\varepsilon$ are the chromophore extinction coefficients, $c$ are the concentrations, and $l$ are the mean photon path length through each of water and the pur\' ee sample. 

Coarse-grained bulk nutrient estimation was accomplished through predicting relative nutritional density using a deep relative nutrient density autoencoder network (for more detail see~\cite{pfisterer2018}). In this study, 400 RGB samples of size $25\times50$ were extracted from the transmittance image acquisitions of the pur\' ee samples. Using the transmittance data, we fine-tuned our reflectance-mode pretrained sweet potato network from~\cite{pfisterer2018} to a maximum of 400 epochs with a combined random 80\% (n=320) of transmittance samples of 6 and 8 month sweet potato to form a general fine-tuned ``sweet potato" relative nutrient density network. To evaluate the accuracy of the proposed deep autoencoder network, we leveraged the remaining 20\% of transmittance samples and compute accuracy for each of: global sweet potato (\eg, 6 month and 8 month combined), and individually for 6 and 8 month sweet potato. Five separate runs were performed.


\section{Results and Discussion}

\begin{figure*}
\begin{center}
\fbox{\includegraphics[width=1.0\linewidth]{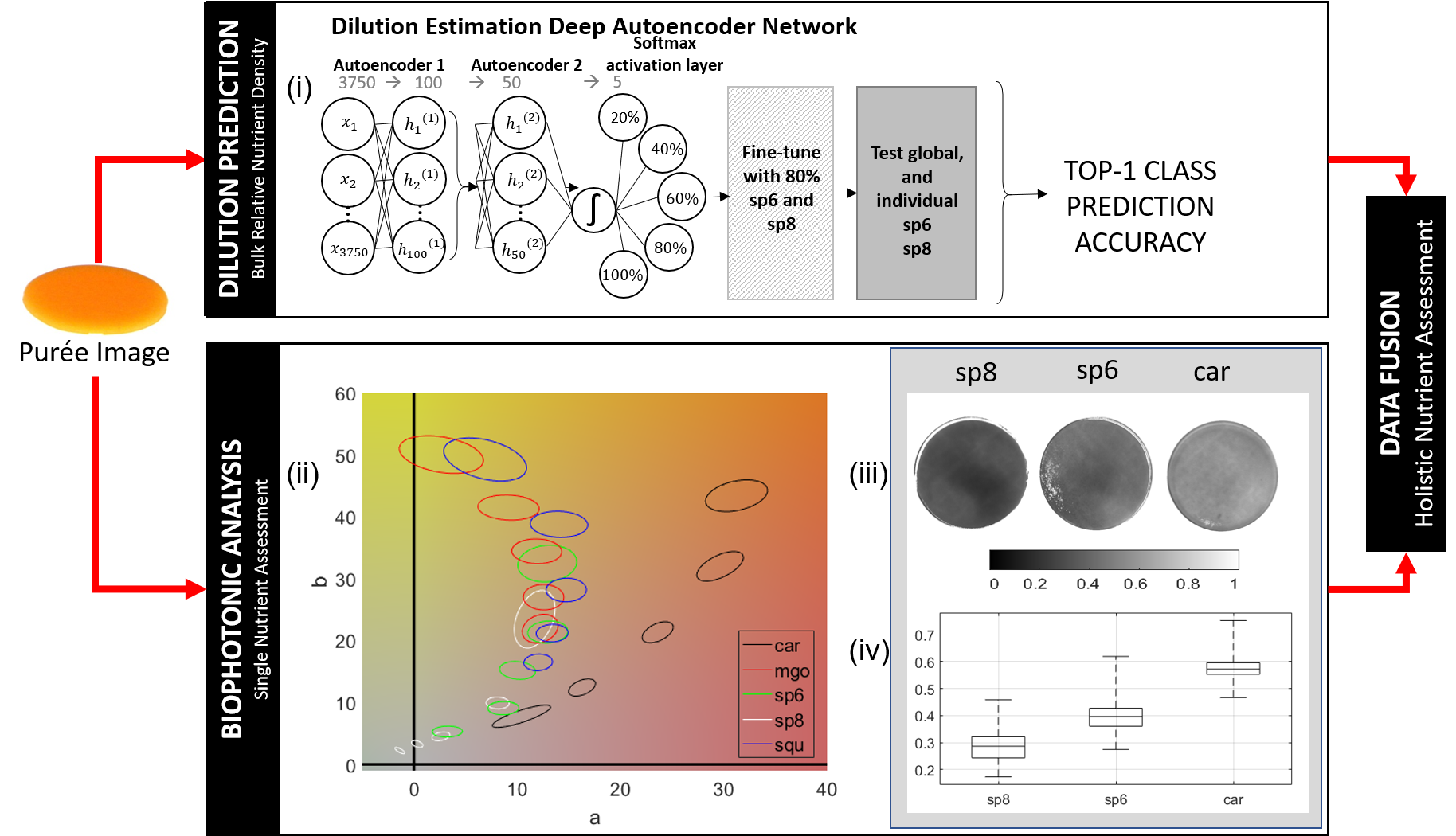}} 
\end{center}
   \caption{Two pronged pur\' ee system analysis comprised of coarse-grained bulk nutrient estimation (top) and single nutrient assessment through biophotonic analysis for fine-grained nutritional assessment (bottom). (i) Deep Autoencoder Network architecture for relative nutrient density estimation. (ii) L-normalized a*b* plots of each puree dilution series in 15 mL samples. \% transmittance maps (iii) and plots (iv) across R channel in highest (sp8 and sp6) and lowest (car) vit A containing samples at 20\% relative dilution.}

\label{fig:Lab}
\end{figure*}

First, we discuss visual and biophotonic trends observed and use this to reinforce our discussion and interpretation of our proposed fine-tuned deep autoencoder network. Analyzing the CIELAB space of the biophotonics transmittance data, we observed a trend observed with higher values of Vitamin A nearer to the origin with lower values arching upward and to the right (Figure~\ref{fig:Lab}, Biophotonic Analysis left pane). This is congruent with the visual appearance of the pur\' ees; within a flavor the more highly concentrated pur\' ee samples were darker and between flavors, there was an observable color difference. With the exception of carrots, the redder or more orange pur\' ees had higher \% Daily Value (\%DV) amounts of vitamin A. One observation was while sweet potato had the highest \%DV of vitamin A ($\%DV_{vitA}$), carrot, with the lowest $\%DV_{vitA}$ appeared more red. Perhaps one reason for this observation is that the optical activity of vitamin A can be further broken down into  contributions from $\beta$- and $\alpha$-carotene. When considered together, it would seem carrot has more carotenoid absorbers present. Since carotenoids absorb in the blue-green range, perhaps a larger relative amount of red spectra are getting transmitted, which may account for the higher \% transmittance in carrot compared to sweet potato as depicted in Figure~\ref{fig:Lab} (iii). An additional observation was that while the nutritional composition for beginner baby food (sp6) and intermediate baby food (sp8) were similar, visually they appeared different; sp6 was slightly redder and lighter than its sp8 counterpart and is shown in Figure~\ref{fig:Lab} (ii) and (iii) implying there were some nutritional difference perhaps not accounted for within the nutritional label.

Given there were visible differences between flavours, we wanted to explore whether it was possible to develop a generalizable deep autoencoder network for predicting relative nutritional density related to vitamin~A concentration. Table~1 shows results using a fine tuned sweet potato deep network. For the combined testing on sp6 and sp8, we achieved a maximum top-1 prediction accuracy of 80\%, with each the accuracy of sp6 and sp8 of 79\%, and 81\%, respectively. One potential contributing factor to errors is the overlap between classes based on visual and CIELAB similarity. For example, in Figure~\ref{fig:Lab} (ii) there are three white ovals near the intercept which correspond to sp8$_{100\%}$, sp8$_{80\%}$, and sp8$_{60\%}$ and these three overlap with the lower most green oval which belongs to sp6$_{100\%}$. With more data for fine-tuning, this approach can be expanded to additional flavours as well given that all except for carrot followed a similar trend but shifted up the arch.

\begin{table}[]
\small
\caption{Summary of sweet potato dilution prediction network accuracy}
\begin{tabular}{llll}
\hline
\textbf{Network Run} & \textbf{sp6+sp8} & \textbf{sp6 only} & \textbf{sp8 only} \\ \hline
1503191536           & 79\%                    & 68\%              & 92\%              \\
1503191538           & 80\%                    & 79\%              & 81\%              \\
1503191540           & 75\%                    & 69\%              & 81\%              \\
1503191542           & 79\%                    & 74\%              & 83\%              \\
1503191544           & 78\%                    & 76\%              & 79\%              \\ \hline
\textbf{Average ($\mu \pm \sigma$)}:             & 78\% $\pm$ 2\%                    & 73\% $\pm$ 5\%              & 83\% $\pm$ 5\%              \\
\textbf{Max}:                 & 80\%                    & 79\%              & 92\%              \\ \hline
\label{tab:networkresults}
\end{tabular}
\vspace{-0.1in}
\end{table}

Moving forward, only considering vitamin A content as contributing to photon absorption events is likely over-simplified especially when additional nutritional constituents may affect incident photon absorption (\eg, chlorophyll, iron) and scattering (\eg, fat, starch) events. Additionally, actual values of vitamin A in the pur\' ee may differ from its raw counterpart due to vitamin A's thermosensitivity and oxidation susceptibility during processing~\cite{knockaert2012carrot}. Exploring computer vision based techniques on how to distinguish between contributions from biophotonic absorbers and scatterers may further enhance our ability to interpret nutritional quantity and quality. Next steps include the integration and data fusion from these two co-processes and expansion to additional nutrients and across a larger sample of food items.


\section{Conclusion}
We presented our two pronged pur\' ee analysis system comprised of biophotonic, single nutrient analysis and bulk nutrient analysis via a deep autonecoder network with a global sweet potato relative dilution prediction accuracy of 80\%. Building on this work provides an opportunity to inform many quality inspection to fraud identification applications. (Funding: Canada Research Chairs; NSERC)


{\small 
\bibliographystyle{ieee}
\bibliography{egbib}
}

\end{document}